# Robust Monocular Flight in Cluttered Outdoor Environments

Shreyansh Daftry, Sam Zeng, Arbaaz Khan, Debadeepta Dey, Narek Melik-Barkhudarov,
J. Andrew Bagnell and Martial Hebert
Robotics Institute, Carnegie Mellon University, USA

*Abstract*—Recently, there have been numerous advances in the development of biologically inspired lightweight Micro Aerial Vehicles (MAVs). While autonomous navigation is fairly straightforward for large UAVs as expensive sensors and monitoring devices can be employed, robust methods for obstacle avoidance remains a challenging task for MAVs which operate at low altitude in cluttered unstructured environments. Due to payload and power constraints, it is necessary for such systems to have autonomous navigation and flight capabilities using mostly passive sensors such as cameras. In this paper, we describe a robust system that enables autonomous navigation of small agile quad-rotors at low altitude through natural forest environments. We present a direct depth estimation approach that is capable of producing accurate, semi-dense depth-maps in real time. Furthermore, a novel wind-resistant control scheme is presented that enables stable way-point tracking even in the presence of strong winds. We demonstrate the performance of our system through extensive experiments on real images and field tests in a cluttered outdoor environment.

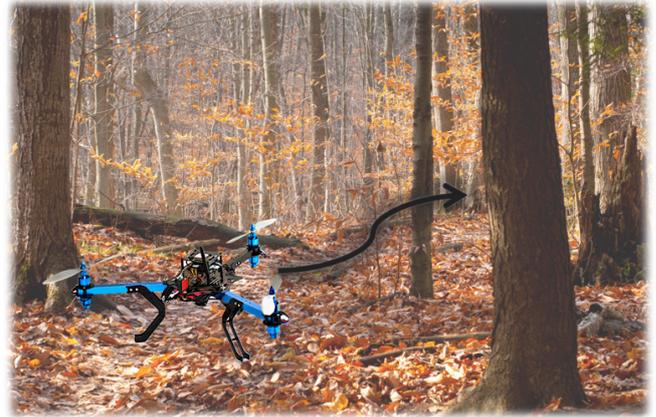

Fig. 1. We present a novel method for high-speed, autonomous MAV flight through dense forest areas, using only passive monocular camera.

## I. Introduction

Flying robots that can quickly navigate through cluttered environments, such as urban canyons and dense forests, have long been an objective of robotics research [28, 37, 48, 25, 39]. Even though nature is home to several species of birds that have mastered the art of maneuverability during high-speed flights, very little have been established so far in realizing the same for Micro Aerial Vehicles (MAVs) [22]. Inspired by this, we take a small step in the direction of building a robust system that allows a MAV to autonomously fly at high speeds of up to 1.5 m/s through a cluttered forest environment, as shown in Figure 1. Our work is primarily concerned with navigating MAVs that have very low payload capabilities, and operate close to the ground where they cannot avoid dense obstacle fields.

In recent years, MAVs have built a formidable résumé by making themselves useful in a number of important applications, from disaster scene surveillance and package delivery to robots used in aerial imaging, architecture and construction. The most important benefit of using such lightweight MAVs is that it allows the capability to fly at high speeds in space-constrained environments. However, in order to function in such unstructured environments with complete autonomy, it is essential that they are able to avoid obstacles and navigate robustly. While autonomous operations and obstacle avoidance of MAVs has been well studied in general, most of these approaches uses laser range finders (lidar) [2, 36] or Kinect cameras (RGB-D sensors) [3]. For agile MAVs with very limited payload and power, it is not feasible to carry such active sensors. Therefore, in this work we are interested in methods for range mapping and obstacle avoidance for agile MAVs through unstructured environments that only have access to a passive monocular camera as its primary sensing modality.

Existing approaches for monocular cluttered flight range from purely reactive approaches [33] that uses an imitation learning based framework for motion prediction to full planning-based approaches [12, 4], that utilize receding horizon control to plan trajectories. While promising results have been shown, several challenges remain that impede truly autonomous flights in general outdoor environments without reliance on unrealistic assumptions. The reactive layer is capable of good obstacle avoidance, however is inherently myopic in nature. This can lead to it being easily stuck in *cul-de-sacs*. In contrast, receding horizon based approaches plan for longer horizons and hence minimize the chances of getting stuck [24]. These approaches require a perception method for accurately constructing the local 3D map of the environment in real-time. However, current perception methods for monocular depth estimation either lack the robustness and flexibility to deal with unconstrained environments or are computationally expensive for real-time robotics applications.

Furthermore, the intuitive difficulties in stability and control of such agile MAVs in the presence of strong winds is often overlooked. An impressive body of research on control and navigation of MAVs has been published recently, including aggressive maneuvers [30], which in principle should be ideal to fly through forests. However, these control schemes have been designed to work under several strong assumptions and controlled laboratory conditions, which are easily alleviated during outdoor cluttered flights. The effect of wind on small MAV flight control can be quite significant, and can lead to fatal conditions when flying in close proximity to clutter or other aerial vehicles [44]. It is the above considerations that motivate our contribution.

Our technical contributions are three-fold: First, we present an end-to-end system consisting of a semi-dense direct visual-odometry based depth estimation approach that is capable of producing accurate depth maps at 15 Hz, even while the MAV is flying forward; is able to track frames even with strong inter-frame rotations using a novel method for visual-inertial fusion on Lie-manifolds, and can robustly deal with uncertainty by making multiple, relevant yet diverse predictions. Secondly, we introduce a novel wind-resistant LQR control based on a learned system model for stable flights in varying wind conditions. Finally, we evaluate our system through extensive experiments and flight tests of more than 2 km in dense forest environments, and show the efficacy of our proposed perception and control modules as compared to the state-of-the-art methods.

## II. RELATED WORK

**Monocular Depth Estimation:** Structure from Motion (SfM) approaches [47, 20, 9] can be used to reconstruct 3D scene geometry from a moving monocular camera. While such SfM approaches are already reasonably fast, they produce sparse maps that are not suited for collision-free navigation: typically the resulting point cloud of 3D features is highly noisy, and more crucially, the absence of visual features in regions with low texture does not necessarily imply free space. In contrast, methods for obtaining dense depth maps [31, 46] are either still too computationally expensive for high-speed flight in a forest or implicitly assume that the camera moves slowly and with sufficient translation motion. On a high speed flight, this is difficult to guarantee, as all flying maneuvers induce a change in attitude which presumably would often lead to a loss of tracking. In recent years, learning based approaches [12, 13] that aim to predict depth directly from visual features have shown promising results. However, these methods are highly susceptible to training data and hence do not generalize to unseen data, making them unsuitable for exploration based navigation tasks.

**Wind Modeling, Detection and Correction:** A significant literature exists on understanding wind characteristics for MAV guidance and navigation. More specifically, work has been done on large scale phenomena such as thermal soaring [1], ridge soaring [26], and exploitation of wind gusts [5, 32, 27]. However for all of these, the environment in which these conditions are studied are relatively benign and free of clutter. Thus, they are strongly reliant on assumptions such as constant wind fields and long time-scale. The reason for such assumptions is that short term, turbulent wind effects are difficult to model even with detailed Computational Fluid Dynamics (CFD) analysis, making it impractical from a control standpoint as studied by Waslander and Wang [44]. In contrast, our work tries to explore the use of simple estimation algorithms to detect current wind disturbances and design a controller that reduces their impact on waypoint tracking accuracy.

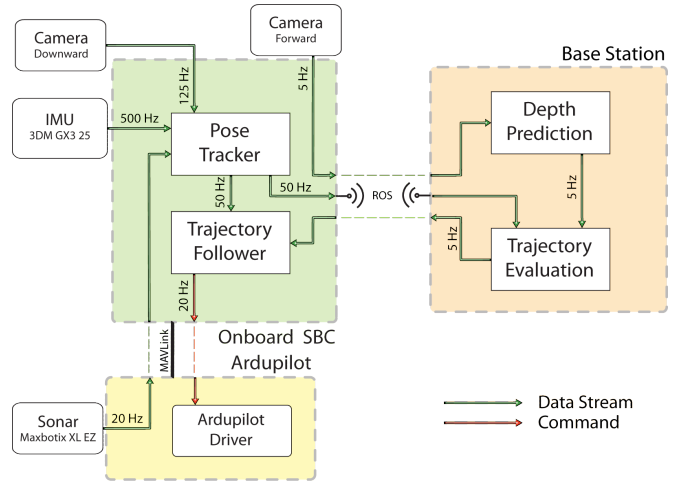

Fig. 2. Schematic diagram of software and hardware modules.

## III. SYSTEM OVERVIEW

We have used a modified version of the 3DR ArduCopter with an onboard quad-core ARM processor and a Microstrain 3DM-GX3-25 IMU. Onboard, there are two monocular cameras: one PlayStation Eye camera facing downward for real time pose estimation and one high-dynamic range PointGrey Chameleon camera for monocular navigation. We have a distributed processing framework as shown in Figure 2, where an image stream from the front facing camera is streamed to the base station where the perception module produces a local 3D scene map; the planning module uses these maps to find the best trajectory to follow and transmits it back to the onboard computer where the control module does trajectory tracking.

## IV. PERCEPTION

The perception system runs on the base station and is responsible for providing a 3D scene structure which can be used for motion planning and control strategies. In the last few years, direct approaches [41, 46, 42, 17] for scene geometry reconstruction have become increasingly popular. Instead of operating solely on visual features, these methods directly work on the image intensities for both mapping and tracking: The world is modeled as a dense surface while in turn new frames are tracked using whole-image alignment. In this section, we build upon recently published work [15, 8] to present

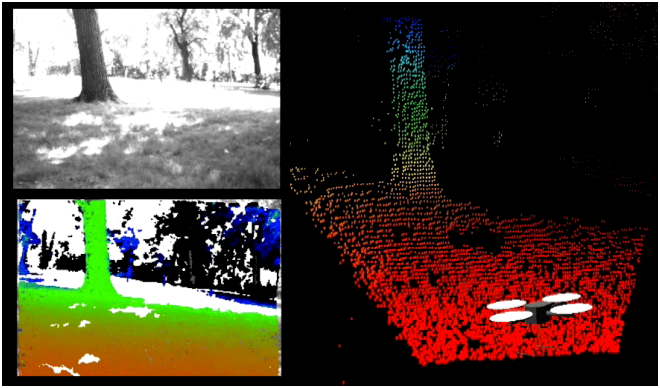

Fig. 3. Semi-dense Depth Map Estimation. (Top left) Example image frame (Bottom left) Depth map. Colormap: Inverted Jet, Black pixels are for invalid pixels (Right) Corresponding 3D point cloud. Color based on height

a complex system focusing on depth map estimation for fast obstacle avoidance and then introduces our contributions for robust behavior in cluttered environment as explained below.

### A. Semi-Dense Monocular Depth Estimation

**Mapping:** The depth estimates are obtained by a probabilistic approach for adaptive-baseline stereo [14]. This method explicitly takes into account the knowledge that in video, small baseline frames occurs before large baseline frames. A subset of pixels is selected for which the disparity is sufficiently large, and for each selected pixel a suitable reference frame is selected and a one dimensional disparity search is performed. The obtained disparity is converted to an inverse-depth representation, where the inverse depth is directly proportional to the disparity. The map is then updated using this inverse depth estimate. The inverse depth map is propagated to subsequent frames, once the pose of the following frames has been determined and refined with new stereo depth measurements. Based on the inverse depth estimate $d_0$ for the pixel, the corresponding 3D point is calculated and projected onto the new frame and assigned to the closest integer pixel position providing the new inverse depth estimate $d_1$. Now, for each frame, after the depth map has been updated, a regularization step is performed by assigning each inverse depth value the average of the surrounding inverse depths, weighted by their respective inverse variance ($\sigma^2$). An example of the obtained depth estimates has been shown in Figure 3. Note: In order to retain sharp edges, which can be critical for detecting trees, we only perform this step if the two adjacent depth values are statistically similar i.e. their variance is within $2\sigma$.

**Tracking:** Given an image $I_M : \Omega \to \mathbf{R}$, we represent the inverse depth map as $D_M : \Omega_D \to \mathbf{R}^+$, where $\Omega_D$ contains all pixels which have a valid depth. The camera pose of the new frame is estimated using direct image alignment. The relative pose $\xi \in SE(3)$ of a new frame $I$, is obtained by directly minimizing the photometric error:

$$E(\xi) := \sum_{x \in \Omega_{D_M}} \|I_M(x) - I(w(x, D_m(x), \xi))\|_\delta \quad (1)$$

where $w : \Omega_{D_M} \times \mathbf{R} \times SE(3) \to \omega$ projects a point $x$ from the reference frame image into the new frame and $\|\cdot\|_\delta$ is the Huber norm to account for outliers. The minimum is computed using iteratively re-weighted Levenberg-Marquardt minimization [15].

**Scale Estimation:** Scale ambiguity is inherent to all monocular visual odometry based methods. This is not critical in visual mapping tasks, where the external scale can be obtained using fiducial markers [34]. However, for obstacle avoidance in real-time, it is required to accurately recover the current scale so that the distance to the object is known in real world units. We resolve the absolute scale $\lambda \in \mathbf{R}^+$ by leveraging motion estimation from a highly accurate single beam laser lite sensor onboard. We measure, at regular intervals (operating at 15 Hz), the estimated distance traveled according to the visual odometry $\mathbf{x}_i \in \mathbf{R}^3$ and the metric sensors $\mathbf{y}_i \in \mathbf{R}^3$. Given such sample pairs $(\mathbf{x}_i, \mathbf{y}_i)$, we obtain a scale $\lambda(t_i) \in \mathbf{R}$ as the running arithmetic average of the quotients $\frac{\|\mathbf{x}_i\|}{\|\mathbf{y}_i\|}$ over a small window size. We further pass the obtained set of scale measurements through a low-pass filter in order to avoid erroneous measurements due to sensor noise. The true scale $\lambda$ thus obtained is used to scale the depth map to real world units.

### B. IMU Pre-Integration on Lie-Manifolds

Visual Odometry (VO) based approaches, like the one described above, are inherently susceptible to strong inter-frame rotations. One possible solution is to use a high frame rate camera. However, this has its own limitations in terms of power and computational complexity. Rather we build upon the recent advancements in visual-inertial navigation methods [45, 21, 38] that use an Inertial Measurement Unit (IMU) to provide robust and accurate inter-frame motion estimates.

We denote with $\mathcal{I}_{ij}$ the set of IMU measurements acquired between two consecutive keyframes $i$ and $j$. We denote with $\mathcal{C}_i$ the camera measurements at keyframe $i$ and $\mathcal{K}_k$ denotes the set of all keyframes up to time k. Usually, each set $\mathcal{I}_{ij}$ contains hundreds of IMU measurements. The set of measurements collected up to time $k$ is

$$\mathcal{Z}_k = \{\mathcal{C}_i, \mathcal{I}_{ij}\}_{i,j \in \mathcal{K}_k} \quad (2)$$

We use this to infer the motion of the system from IMU measurements. All measurements between two keyframes at times $k = i$ and $k = j$ can be summarized in a single compound measurement, named *preintegrated IMU measurement*, which constrains the motion between consecutive keyframes. This concept was first proposed by Lupton and Sukkarieh [29] using Euler angles and Forster et al. [16] extended it, by developing a suitable theory for preintegration on Lie-manifolds. The resulting *preintegrated rotation increment* is given as:

$$\Delta \tilde{R}_{ij} \doteq \sum_{k=i}^{j-1} \mathrm{Exp}((\tilde{w}_k - b_i)\Delta t) \quad (3)$$

where $\tilde{w}_k \in \mathrm{R}^3$ is the instantaneous angular velocity of the $k^{th}$ frame, relative the world frame. $b_i$ is a constant noise bias at time $t_i$ and the shorthand $\Delta t \doteq \sum_{k=i}^{j} \Delta t$. Equation 3 provides an estimate of the rotational motion between time $t_i$ and $t_j$, as estimated from inertial measurements.

Now, the warp function $w$ as described in Eq. 1 can be written as:

$$w(x, D_M, T) = \pi(T\pi^{-1}(x, D_M(x))). \quad (4)$$

where $\pi^{-1}(x, D_M)$ represents the inverse projection function and T $\in SE(3)$ represents the 3D rigid body transformation. The estimated motion transform is updated using the calculated preintegrated inertial rotation from the previous step such that

$$T_{imu} = \begin{bmatrix} \Delta \tilde{R}_{imu} & \mathbf{t} \\ 0 & 1 \end{bmatrix} \quad (5)$$

where $\Delta \tilde{R}_{imu} \in SO(3)$ and $\mathbf{t} \in \mathrm{R}^3$. During optimization, a minimal representation for the pose is required, which is given by the corresponding element $\xi \in SE(3)$ of the associated Lie-algebra such that $\xi_{imu}^{(n)} = \log_{SE(3)}(T_{imu})$. In the prediction step, the new estimate is obtained by multiplication with the computed update:

$$\xi^{(n+1)} = \delta\xi^{(n)} \circ \xi_{imu}^{(n)} \quad (6)$$

where $\delta(\xi)^n$ is computed by solving for the minimum of a Gauss-Newton second order approximation of the photometric error $E$. As the optimizer is fed with more robust rotational estimates, the resulting pose after visual-inertial fusion is less susceptible to strong inter-frame rotations.

*C. Multiple Predictions*

The monocular depth estimates are often noisy and inaccurate due to the challenging nature of the problem. A planning system must incorporate this uncertainty to achieve safe flight. Figure 4 illustrates the difficulty of trying to train a predictive method for building a perception system for collision avoidance. Figure 4 (left) shows a ground truth location of trees in the vicinity of an autonomous UAV. Figure 4 (middle) shows the location of the trees as predicted by the perception system. In this prediction the trees on the left and far away in front are predicted correctly but the tree on the right is predicted close to the UAV. This will cause the UAV to dodge a ghost obstacle. While this is bad, it is not fatal because the UAV will not crash but make some extraneous motions. But the prediction of trees in Figure 4 (right) is potentially fatal. Here the trees far away in front and on the right are correctly predicted whereas the tree on the left originally close to the UAV, is mispredicted to be far away. This type of mistake will cause the UAV to crash into an obstacle it does not know is there.

Ideally, a vision-based perception system should be trained to minimize loss functions which will penalize such fatal predictions more than other kind of predictions. But even writing down such a loss function is difficult. Therefore most monocular depth perception systems try to minimize easy to optimize surrogate loss functions like regularized $L_1$ or $L_2$ loss [35]. We try to reduce the probability of collision by generating multiple interpretations of the scene to hedge against the risk of committing to a single potentially fatal interpretation. Specifically we generate 3 interpretations of the scene and evaluate the trajectories in all of them. The trajectory which is least likely to collide on average in all interpretations is then chosen as the one to traverse.

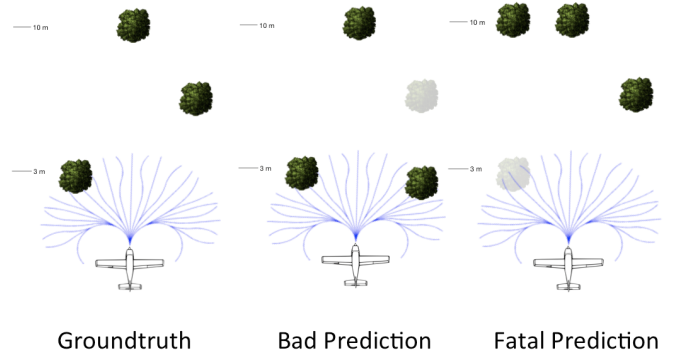

Fig. 4. Illustration of the complicated nature of the loss function for collision avoidance. (Left) Groundtruth tree locations (Middle) Bad prediction where a tree is predicted closer than it actually is located (Right) Fatal prediction where a tree close by is mispredicted further away.

In previous work, Dey et al. [11] have developed techniques for predicting a budgeted number of interpretations of an environment with applications to manipulation, planning and control. These approaches try to come up with a small number of relevant but diverse interpretations of the scene so that at least one of them is correct. In a similar spirit, we make multiple predictions by utilizing the variance of the estimated inverse depth which is already calculated in our direct visual odometry framework. At every pixel, the variance of the inverse depth is used to find the inverse depth value one standard deviation away from the mean (both lower than and higher than the mean value) and inverted to obtain a depth value. So, a total of 3 depth predictions are made: 1) mean depth estimate 2) depth estimate at one standard deviation lower than the mean depth at every pixel and 3) depth estimate at one standard deviation greater than the mean depth at every pixel.

## V. PLANNING AND CONTROL

*A. Receding Horizon Control*

We use a semi-global planning approach in a receding horizon control scheme [23]. Once the planner module receives a scaled-depth map from the perception module, the local point cloud is updated using the current pose of the MAV. A trajectory library of 78 trajectories of length 5 meters is budgeted and picked from a much larger library of 2401 trajectories using the maximum dispersion algorithm by Green et al. [18]. Further, for each of the budgeted trajectories a score value for every point in the point cloud, is calculated by taking into account several factors, including the distance to goal, cost of collision, etc., and the optimal trajectory to follow is selected. The control module takes as input the selected trajectory to follow and generates waypoints to track using a pure pursuit strategy [6]. Using the pose estimates of the vehicle, the MAV moves towards the next waypoint using a LQR controller as described in Section V-C. For further details on our planning approach, we direct the reader to previous work by Dey et al. [12].

*B. State Estimation*

In receding horizon, one only needs a relative, consistent pose estimation system as trajectories are followed only for a short duration. As looking forward to determine pose is ill-conditioned due to a lack of parallax, we use a downward looking camera in conjunction with a sonar for determining this relative pose. We used a variant of a simple algorithm that has been presented quite often, most recently in [19]. This approach uses a Kanade-Lucas-Tomasi (KLT) tracker [43] to detect where each pixel in a grid of pixels moves over consecutive frames, and estimating the mean flow from these after rejecting outliers. Additionally, out-of-plane camera ego-motion is compensated using motion information from the IMU and the metric scale is estimated from sonar. The computed instantaneous relative velocity between the camera and ground is integrated over time to get position.

*C. Wind-Resistant LQR Control*

In this section, we describe our proposed approach for wind-resistant LQR control. The problem under consideration can be described as follows: Given the knowledge of the current and past states of the MAV, can we estimate the current wind conditions and determine appropriate control actions that would help recover from these expected disturbances? To this end, we propose to learn a dynamic model of our MAV and observe the errors in the model's predictions in windy conditions, and compare them to errors in calm conditions in order to determine the magnitude and direction of the wind. These observations can be built into a learned model for wind behavior, which allows for future control commands to be adjusted for the predicted effects of the wind.

**System Identification:** Given the application and the requisite desired performance, we are only interested in the vehicle dynamics involving small angular deviations and angular velocities. Thus, we linearize the dynamics and approximate the MAV as a linear system. Now, while this is trivial for roll and pitch, yaw is not limited to small angles in our motion model, resulting in a non-linear behavior. This non-linearity can be removed by building our model in the world frame and transforming the results to the robot frame when necessary. Thus, the robot's motion model can be represented by the following state space equation:

$$x(t+1) = \mathcal{A}x(t) + \mathcal{B}u(t) \quad (7)$$

where the state vector $x = [\text{x}, \text{y}, \dot{\text{x}}, \dot{\text{y}}, \theta]$ consists of x and y positions, x and y velocities and the yaw angle and $u = [\tilde{\eta}, \tilde{\phi}, \tilde{\theta}]$ is the control input vector consisting of roll, pitch and yaw control commands. The matrices $\mathcal{A}$ and $\mathcal{B}$ can be obtained analytically by system parameters such as mass, moment of inertia, center of mass, etc. However, these parameters are difficult to obtain in practice. Moreover, this approach cannot model the effects of the time varying aerodynamic forces. Alternatively, we run a large number of experiments to record the state and command data from actual flight and then learn the matrices $\mathcal{A}$ and $\mathcal{B}$ that best fit the data. In this work, we recast the system identification problem into a least square solution to an overdetermined set of linear equations. Specifically, the state space equations can be condensed into the form

$$X(t+1) = [\mathcal{A}, \mathcal{B}] \begin{bmatrix} x(t) \\ u(t) \end{bmatrix} \quad (8)$$

where the matrix $[\mathcal{A}, \mathcal{B}]$ can be solved using linear regression.

**Wind Modeling, Detection, and Resistance:** Traditional approaches for wind detection, model the gust forces assuming a static dominant direction derived from the Drysden model [40]. However, this model is only applicable to MAVs in open terrain and fails for MAV flights in cluttered environments like forests, as the gush of wind has a profile of varying velocity on a time-scale shorter than the scale of the vehicle flight time. Thus, we use an alternate approach for determining the wind disturbance, which relies on the use of the acceleration data. In our proposed approach we attempt to understand the behaviors of our learned system model in both windy and wind-free conditions. Any deviations that should occur, given the current state and inputs, can be attributed to wind disturbances. In particular, the effects of the wind on the system dynamics can be observed by analyzing the differences in model errors in both conditions (See Figure 8a). The magnitude of the difference at a given time represents the current wind impact on the MAV and the direction can be determined by x and y components of the velocity error since the wind should cause unexpected acceleration in the direction of the wind.

Furthermore, the differences in model errors between the average error distribution and the current real time distribution at a given time can be broken down into each state's error. Additionally, if we assume that the current wind magnitude and direction is the same as it was in the previous time step, the MAV model's state can be extended to include wind resulting in a new model:

$$x(t+1) = \mathcal{A}x(t) + \mathcal{B}u(t) + w(t) \quad (9)$$

where $w = [w_x, w_y, w_{\dot{x}}, w_{\dot{y}}, 0]$ are the wind-bias correction terms corresponding to x, y positions and x, y velocities. We assume the wind to have no impact on the yaw angles. The validity of these assumptions can be determined by observing the variance in the previous $N$ errors. If the variance is low, it is likely that the next error will be the same as the past few and so the model should be updated to correct for this error since wind comes in low frequency bursts. If the variance is high, then the error is can be attributed to random noise or other modeling error and it is ignored in future predictions. This process can be executed in real time during flight allowing for current estimates for the wind force.

**LQR Controller Design:** The system dynamics model is used to implement an output feedback control law to track a given trajectory. An LQR controller [49], $u = -Kx$ was designed to minimize the following quadratic cost function

$$J = \lim_{N \to \infty} E(\sum_{k=1}^{N} x(t)^T \mathcal{Q} x(t) + u(t)^T \mathcal{R} u(t)) \qquad (10)$$

where $\mathcal{Q} \geq 0$ and $\mathcal{R} \geq 0$ are the weighing matrices that reflect the tradeoff between regulation performance and control effort. The diagonal entries in the weighing matrices are iteratively tuned [10] to ensure a good transient response without saturating the control inputs.

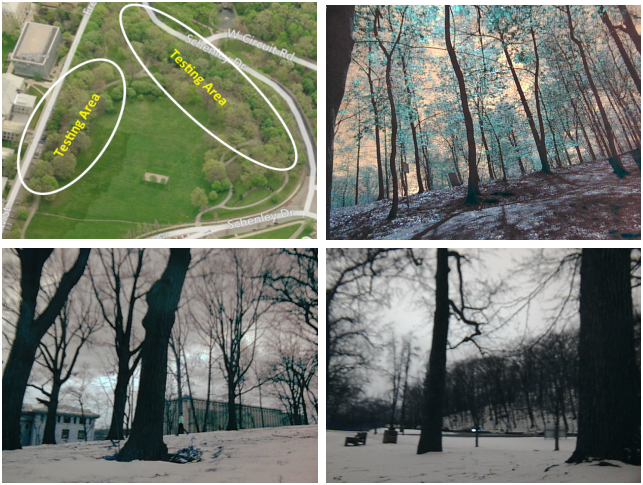

Fig. 5. (Top Left) Testing area near Carnegie Mellon University, Pittsburgh, USA and (Right) An example image illustrating the density of the forest environment during summer. (Below) Example images during winter conditions

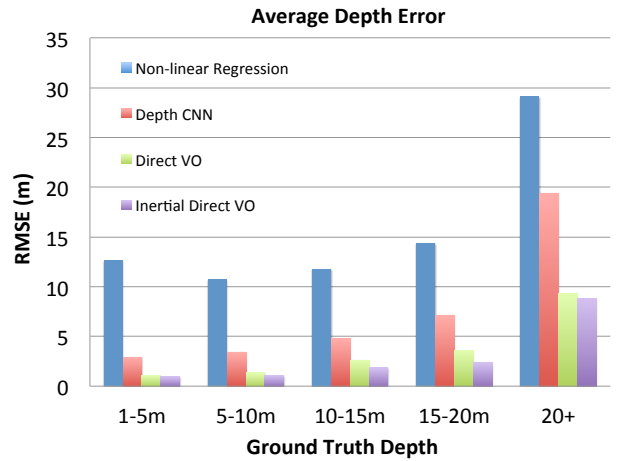

Fig. 7. Average root-mean-squared-error (RMSE) binned over groundtruth depth buckets of [1, 5] m, [5, 10] m, etc. Groundtruth depth images are obtained from stereo image processing.

## VI. EXPERIMENTS

In this section we analyze the performance of our proposed method for robust monocular MAV flight through outdoor cluttered environments. All the experiments were conducted in a densely cluttered forest area as shown in Figure 5, while restraining the MAV through a light-weight tether. It is to be noted that the tether is only for compliance to federal regulations and does not limit the feasibility of a free flight.

### A. Performance Evaluation of Perception Module

**Ground truth.** In order to collect groundtruth depth values, a Bumblebee color stereo camera pair ($1024 \times 768$ at $20$ Hz) was rigidly mounted with respect to the front camera using a custom 3D printed fiber plastic encasing. We calibrate the rigid body transform between the front camera and the left camera of the stereo pair using the flexible camera calibration toolbox from Daftry et al. [7]. Stereo depth images and front camera images are recorded simultaneously while flying the drone. The depth images are then transformed to the front camera's coordinate system to provide groundtruth depth values for every pixel. The corpus of train and test data consists of $50$k and $10$k images respectively, and was acquired from the same test site.

**Accuracy compared to Ground truth.** We evaluate performance of the perception module for depth map accuracy. Figure 7 shows the average depth error against ground truth depth images obtained from stereo processing as explained above. As baseline, we compare our method to state-of-the-art approaches for real-time monocular depth estimation using non-linear regression [12] and Depth CNN [13]. The error plots have been binned with respect to the ground-truth, in order to show the quality of depth maps at different working scale (For a receding horizon scheme, accurate reconstruction of objects only upto a $10$ m range is relevant). It can be observed that direct visual odometry performs really well with low error values up to [15, 20] m. Please note that both the baseline methods are learning-based approaches and thus have been trained on the training data collected. This graph nevertheless serves to show the accuracy of the our proposed perception method.

**Benefits of Visual-Inertial Fusion.** As discussed previously, tracking-based methods are highly susceptible to strong inter-frame rotations. We can observe from Figure. 7 that fusion of inertial data according to out proposed formulation improves the accuracy of the eventual depth maps. In particular, this can be directly attributed to two reasons: First, better initial conditions to the optimizer leads to faster convergence and prevents it from being stuck in a local optima. Secondly, lesser number of tracking losses; each time tracking is lost, the depth maps are initialized to random values and hence results in poor depth maps for the initial few frames till the VO recovers. Quantitatively, we observed the total tracking loss was reduced by 55% in average over a distance of $1$ km. Qualitatively, the estimated depth maps can be observed to be within the uncertainty range of stereo, as shown in Figure 6. Its interesting to note from the qualitative results that our method is able to robustly handle even poor quality images that suffer from over- or under-exposure, shadows, etc.

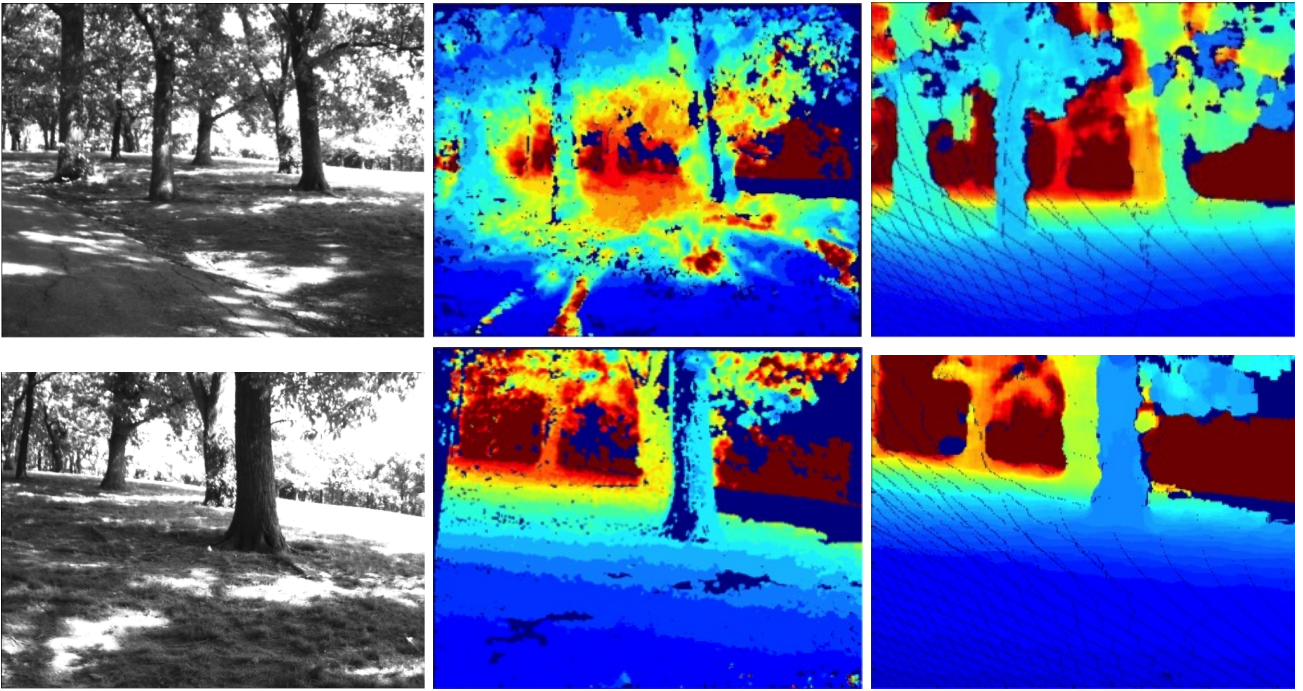

Fig. 6. Qualitative analysis of depth maps produced by our proposed depth estimation approach with respect to stereo ground-truth. (Left) Image (Middle) estimated depth map and (Right) ground-truth depth map. Note: Jet colormap

*B. Wind Estimation and Resistance*

To evaluate the performance of the wind correction, we compared the accuracy of the original model and the adaptive wind correction on state and command data collected during flight under windy conditions. Figure 8b demonstrates how the wind correction can adjust the model's prediction during windy conditions. Specifically, the mean error in y velocity estimation during windy conditions was significantly shifted away from zero due to the force of the wind. Additionally, after wind correction, the mean modeling error returns to being centered around zero. Furthermore, our experiment demonstrated that overall, this wind correction performed significantly better than the original model during windy conditions. To elaborate, figure 8c shows the resulting difference between the wind corrected error and the original model's error during windy flight conditions sorted by their magnitude. As depicted by the magnitude and number of negative errors, the wind corrected model made better predictions then the original model on about 70% of the test data. Overall the wind corrected model demonstrated solid improvement across all elements of the state during flight in windy conditions.

*C. System Performance Evaluation*

Quantitatively, we evaluate the performance of our system by observing the average distance flown autonomously by the MAV over several runs (at 1.5 m/s), before an intervention. An intervention, in this context, is defined as the point at which the pilot needs to overwrite the commands generated by our control system so as to prevent the drone from an inevitable crash. Experiments were performed using both the proposed multiple prediction approach and single best prediction, and results comparing to previous state-of-the-art approaches on monocular reactive control [33] and receding horizon control [12] has been shown in Figure 9. Tests were performed in regions of low and high clutter density (approx. 1 tree per $6 \times 6\ m^2$ and $12 \times 12\ m^2$, respectively).

It can be observed that using our perception approach MAVs can fly, at an average, 6 times further than pure reactive control in high density regions. While this is not surprising given the accuracy of the depth maps produced, it affirms our claim that even when we are moving forward, which is the direction of least parallax, good depth maps can be realized. Moreover, multiple predictions gives a significant boost (upto 78% improvement in low density regions) to the average flight distance over corresponding single prediction approach. In particular, the MAV was able to fly autonomously without crashing over a 580 m distance at average. This validates our intuition from Section IV-C that by avoiding a small number of extra ghost obstacles, we can significantly reduce crashes due to uncertainty.

## VII. CONCLUSION

In this paper, we have presented a robust approach for high-speed, autonomous MAV flight through dense forest environments. Our direct depth estimation approach enables real-time computation of accurate dense depth maps even when the MAV is flying forward, is robust to pure rotations and implicitly handles uncertainty through multiple predictions. Further, we also propose a novel formulation for wind-resistant control that allows stable waypoint tracking in the presence of strong gusts of wind. During a significant amount of outdoor experiments with flights over a distance of $2$ km, our approach has avoided more than 530 trees in environments of varying density. In future work, we hope to move towards complete onboard computing of all modules to reduce latency. Another central future effort is to integrate the purely reactive approach with the deliberative scheme detailed here, for better performance. This will allow us to perform even longer flights, in even denser forests and other cluttered environments.

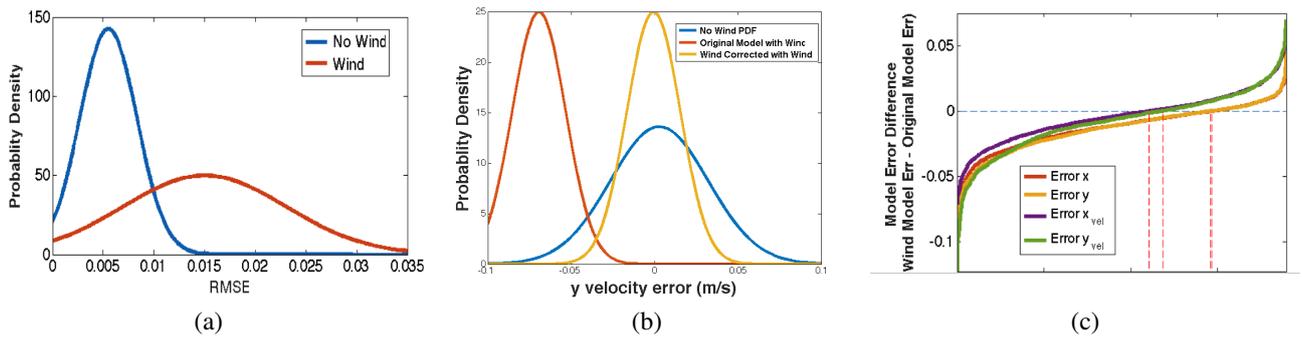

Fig. 8. Wind Modeling, Detection and Correction. (a) Analysis of the differences in model prediction errors can be used to estimate the effects of the wind on the MAV. (b) Example of wind correction using our learned system model (c) Difference between the wind corrected error and the original model's error during windy flight show that wind corrected model made better predictions.

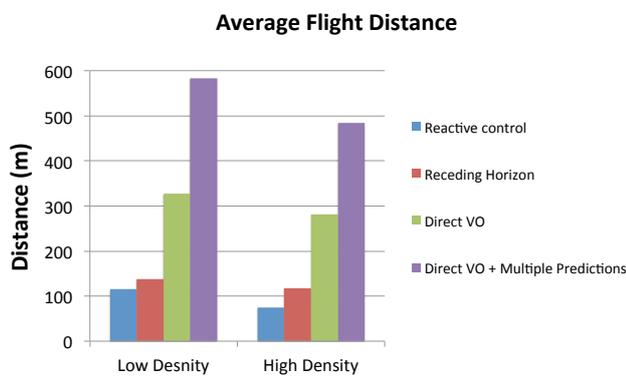

Fig. 9. Average flight distance per intervention for (a) Low and (b) High density regions. For both, the corresponding multiple prediction variant performs significantly better.